\documentclass[journal]{IEEEtran}
\ifCLASSINFOpdf
  \usepackage[pdftex]{graphicx}
\else
\fi
\usepackage{subfigure}% Actually replaced by `subfig`!
\usepackage{amsmath}
\usepackage{dblfloatfix}

\begin{document}
\title{Multi-Task Learning for Segmentation of Building Footprints with Deep Neural Networks}
%
%
% author names and IEEE memberships
% note positions of commas and nonbreaking spaces ( ~ ) LaTeX will not break
% a structure at a ~ so this keeps an author's name from being broken across
% two lines.
% use \thanks{} to gain access to the first footnote area
% a separate \thanks must be used for each paragraph as LaTeX2e's \thanks
% was not built to handle multiple paragraphs
%

%\author{Benjamin~Bischkel,~Patrick~Helber,~Joachim~Folz,~Damian~Borth,~Andreas~Dengel}

\author{
	\fontsize{12pt}{0}\fontfamily{phv}\selectfont 
	~ Benjamin Bischke~\textsuperscript{1, 2}
	~ Patrick Helber~\textsuperscript{1, 2}
	~ Joachim Folz~\textsuperscript{2}
	~ Damian Borth~\textsuperscript{2} ~  
	~ Andreas Dengel~\textsuperscript{1, 2} \\
	\fontsize{10pt}{0}\fontfamily{phv}\selectfont 
	\hspace{-0.25em}
	$^1$University of Kaiserslautern, Germany\\ ~~~~ $^2$German Research Center for Artificial Intelligence (DFKI), Germany  \\ 
	\fontsize{10pt}{0}\fontfamily{phv}\selectfont 
	\{Benjamin.Bischke, Patrick.Helber, Joachim.Folz, Damian.Borth, Andreas.Dengel\}@dfki.de  \\ 
}
\maketitle

% As a general rule, do not put math, special symbols or citations
% in the abstract or keywords.
\begin{abstract}
The increased availability of high resolution satellite imagery allows to sense very detailed structures on the surface of our planet. Access to such information opens up new directions in the analysis of remote sensing imagery. However, at the same time this raises a set of new challenges for existing pixel-based prediction methods, such as semantic segmentation approaches. While deep neural networks have achieved significant advances in the semantic segmentation of high resolution images in the past, most of the existing approaches tend to produce predictions with poor boundaries. In this paper, we address the problem of preserving semantic segmentation boundaries in high resolution satellite imagery by introducing a new cascaded multi-task loss. We evaluate our approach on Inria Aerial Image Labeling Dataset which contains large-scale and high resolution images. Our results show that we are able to outperform state-of-the-art methods by 8.3\% without any additional post-processing step.
\end{abstract}

% Note that keywords are not normally used for peerreview papers.

% For peer review papers, you can put extra information on the cover
% page as needed:
% \ifCLASSOPTIONpeerreview
% \begin{center} \bfseries EDICS Category: 3-BBND \end{center}
% \fi
%
% For peerreview papers, this IEEEtran command inserts a page break and
% creates the second title. It will be ignored for other modes.
\IEEEpeerreviewmaketitle

%%%%%%%%%%%%%%%%%%%%%%%%%
%%
%%
%%		Introduction
%%
%%
%%%%%%%%%%%%%%%%%%%%%%%%%

\section{Introduction}
\IEEEPARstart{T}{he} increasing number of satellites constantly sensing our planet has led to a tremendous amount of data being collected. Recently released datasets such as the EuroSAT \cite{helber2017eurosat} and Inria Building Dataset \cite{maggiori2017can} contain images which cover a large surface of our earth including numerous cities. Today, labels employed for land-use classification and points-of-interest detection such as roads, buildings, agriculture areas, forests, etc. are primarily annotated manually. Building upon the recent advances in deep learning, we show how automated approaches can be used to support and reduce such a laborious labeling effort.

In this paper, we focus on the segmentation of building footprints from high resolution satellite imagery. This is of vital importance numerous domains such as Urban Planning, Sociology, and Emergency Response (as observed by the necessity to map buildings in Haiti during the response to hurricane 'Matthew' in 2016). The task of automatically segmenting building footprints at a global scale is a challenging task since satellite images often contain deviations depending up on the geographic location. Such deviations are caused by different urban settlements which can be densely or sparsely populated, having different shapes of buildings and varying illuminations due to local atmospheric distortions.

To address the problem of the global variation, Maggiori et. al. \cite{maggiori2017can} created a benchmark database of labeled imagery covering multiple urban landscapes, ranging from highly dense metropolitan financial districts to alpine resorts. The authors observed that the shape of the building predictions on high resolution images is often rounded and does not exhibit straight boundaries which buildings usually have (see Fig. \ref{fig:rounded_predictions}).

The problem of ``blobby" predictions has been addressed in the work of Maggiori et. al.  \cite{maggiori2016high, maggiori2017can} where they suggested to train an MLP on top of an FCN to improve the segmentation prediction. Similarly, Marmanis et. al. \cite{marmanis2016classification} proposed to combine feature maps from multiple networks at different scales and make the final predictions on top of these concatenated feature maps. One drawback of such approaches is that the model complexity and number of parameters is increased by the different networks. Furthermore, it leads to a very high memory consumption since feature maps at each resolution of the network have to be up-sampled to the full size of the output image.

%One problem with these approaches is that it can often not be easily learned in an end-to-end fashion due to the separate networks. Additionally, methods in 

\begin{figure}[tb]
		\centerline{\includegraphics[width=2.8cm]{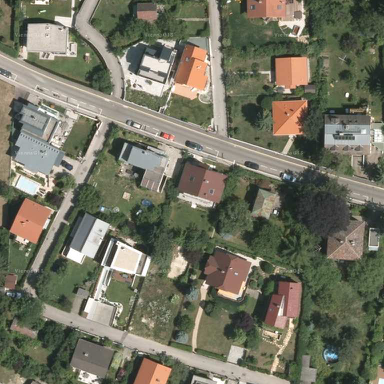}
			\includegraphics[width=2.8cm]{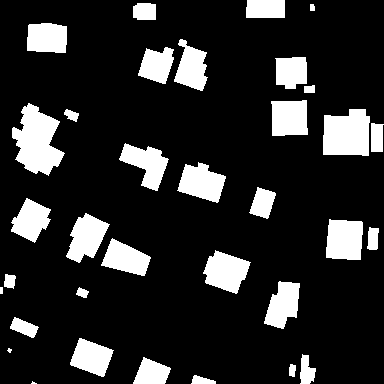}
			\includegraphics[width=2.8cm]{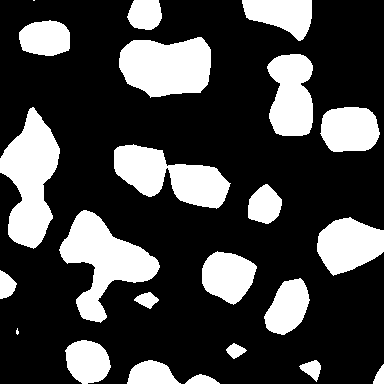}
		}
\end{figure}
\begin{figure}
			\vspace{-11.pt}
	\centering
	\centerline{\includegraphics[width=2.8cm]{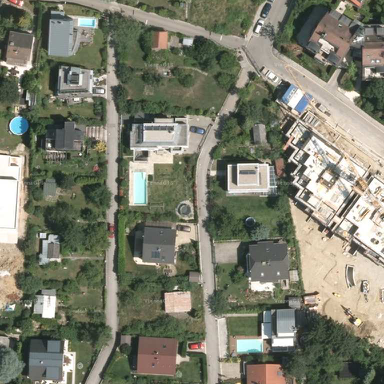}
	\includegraphics[width=2.8cm]{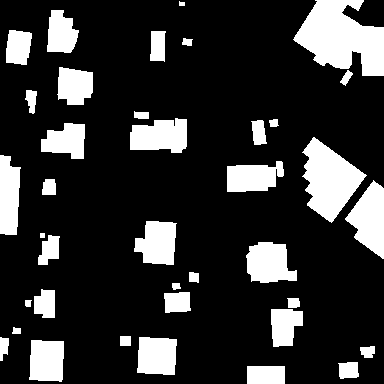}
	\includegraphics[width=2.8cm]{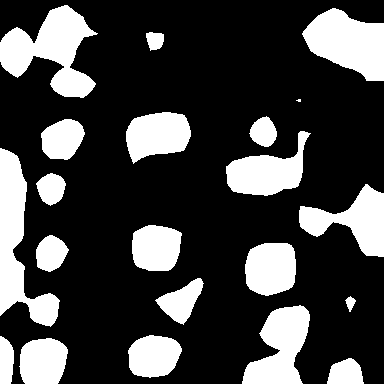}
}
	\caption{Two examples of RGB satellite image (left), ground truth masks for building footprints (middle), and corresponding predictions by a FCN network \cite{shelhamer2017fully} (right). It can be seen that the prediction overlap well with the ground truth but often fail to reflect the boundaries which are present in the ground truth masks.}
	\label{fig:rounded_predictions}
\end{figure}

In this paper, we propose a cascaded multi-task loss along with using a deeper network architecture (than used in \cite{maggiori2017can, maggiori2016high}), to overcome the problem of ``blobby" predictions. Our approach incorporates boundary information of buildings and improves the segmentation results while using less memory at inference time compared to \cite{maggiori2017can}. In this regard, the contributions of this paper can be summarized as follows:
\begin{itemize}
		\item Focusing on the Inria Aerial Image Labeling Dataset, we perform detailed experiments with VGG16 as encoder for segmentation networks. We show that features learned by VGG16 are more powerful than the ones extracted from networks proposed in \cite{maggiori2016high, maggiori2017can}. We additionally evaluate the importance of different decoder architectures.
		\item We introduce an uncertainty weighted and cascaded multi-task loss based on distance transform to improve semantic segmentation predictions of deep neural networks. Thereby, we achieved increased accuracy for the segmentation of building footprints in remote sensing imagery. Learning with our multi-task loss, we are able to improve the performance of the network by 3.1\% without any major changes in the architecture.
		\item We show that our approach outperforms current state-of-the-art accuracy for the Intersection over Union (IoU) on the validation set significantly by 8.3\% without any post-processing step.
\end{itemize}
%- rather than learning different represnetaiotns in different networks and fusing them for the final label prediction, our goal is to let the network learn the shared representation for boundary and semantic features.

%%%%%%%%%%%%%%%%%%%%%%%%%
%%
%%
%%		Related Work	
%%
%%
%%%%%%%%%%%%%%%%%%%%%%%%%
\section{Related Work}

Semantic Segmentation is one of the core challenges in computer vision. Convolutional neural networks such as fully convolutional networks or encoder-decoder based architectures have been successfully applied to this problem and outperformed traditional computer vision approaches marginally. A detailed survey of deep neural network based architectures for semantic segmentation can be found in \cite{garcia2017review}.
One of the main problems when applying CNNs on semantic segmentation tasks is the down-sampling with pooling layers. This increases the field of view of convolutional kernels but loses at the same time high-frequency details in the image. Past work has addressed this issue by reintroducing high frequency details via skip-connections \cite{shelhamer2017fully, badrinarayanan2015segnet, ronneberger2015u}, dilated convolutions \cite{yu2015multi, zhao2016pyramid, chen2016deeplab, chen2017rethinking} and expensive post-processing with conditional random fields (CRF's) \cite{yu2015multi, chen2016deeplab, lin2016refinenet}. While these approaches are able to improve the overall segmentation results, the boundaries between two different semantic classes can often not be well segmented. The importance of segmenting correct semantic boundaries is also considered in recent segmentation datasets such as the \textit{MIT Places Challenge 2017} \cite{zhou2017scene} which evaluates besides the predicted segmentation masks also the accuracy of semantic boundary predictions. Furthermore, it can be observed that more recent network architectures focus on the incorporation of boundary information in the models. This is often achieved on the architecture level by introducing special boundary refinement modules \cite{lin2016refinenet, peng2017large}, on the fusion level by combining feature maps with boundary predictions \cite{marmanis2016classification} or by using a different output representation in the training \cite{hayder2017boundary, yuan2016automatic}. Closest to our work are the approach of Hayder et. al. \cite{hayder2017boundary} and Yuan \cite{yuan2016automatic} which train the network to predict distance classes to object boundaries instead of a segmentation mask. Our work differs from this work, that we additionally predict semantic labels through a multi-task loss and further improve the segmentation results. 

%Our evaluation shows that the cascading of distance and boundary layer  this improves the results.

In context of remote sensing applications the extraction of building footprints has been extensively studied in the past decade. The problem has be addressed by traditional computer vision techniques using hand-crafted features such as vegetation indices \cite{sun2017high, wei2012adaboost}, texture and color features \cite{jabari2014stereo, sun2017high, wei2012adaboost} along with traditional machine learning classifiers (e.g., AdaBoost, support vector machines (SVM), random forests (RF)). Often an additional post-processing step is applied to refine the segmentation results \cite{sun2017high, niemeyer2013classification}.
More recent work uses pixel level convolutional neural networks for building detection. Zhang et. al. \cite{zhang2016cnn} trained a CNN on Google Earth images and applied an additional post processing step using maximum suppression to remove false buildings. Yuan \cite{yuan2016automatic} used a FCN to predict the pixel distance to boundaries and thresholded the predictions to get the final segmentation mask. One of the first approaches which does not rely on an additional post-processing step was proposed by Huang et. al \cite{huang2016building}. They trained a deconvolutional network with two parallel streams on RGB and NRG band combinations and fused the predictions of the two streams. A similar two stream network was used by Marmanis et. al. \cite{marmanis2016classification} which processed RGB and DEM information in parallel. Similar to our approach, the focus of Marmanis's et. al. \cite{marmanis2016classification} work is to preserve boundary information on segmentation classes. This was achieved by first using SegNet as feature extractor (as in our work) and applying additionally an edge detection network (HED) to extract edge information. The boundary predictions are injected into the network by concatenating the feature maps of SegNet with the edge prediction. Our work is different from this approach, that we do not want to extract boundary and semantic information by two different networks and fuse this information at later stages. Our goal is to rather train a single network such that a shared representation for boundary and segmentation prediction can be learned. This reduces the overall complexity of the model and avoids problems such as class-agnostic edge predictions. Recently, Maggiori et. al. \cite{maggiori2017can} showed that previously trained CNN's based on the Massachusetts dataset, generalize poorly to satellite images taken from other cities. Therefore, they released a new large-scale dataset containing high-resolution satellite images. We evaluate our method on this new dataset and compare the results against the best performing methods.

\begin{figure*}[ht!]
	\centering
	\centerline{
		\subfigure[]{\includegraphics[width=3.7cm]{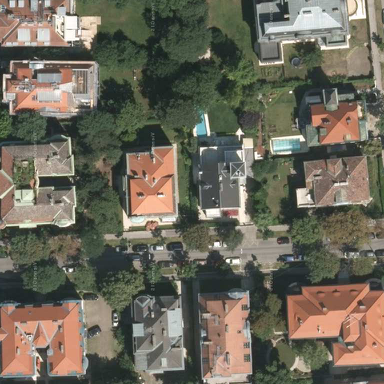}}
		\subfigure[]{\includegraphics[width=3.7cm]{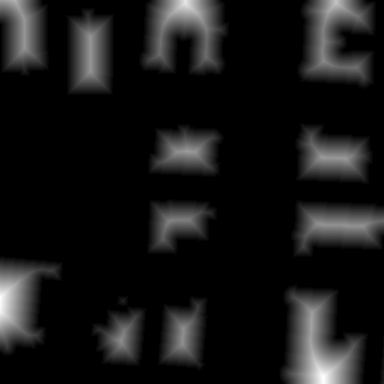}}
		\subfigure[]{\includegraphics[width=3.7cm]{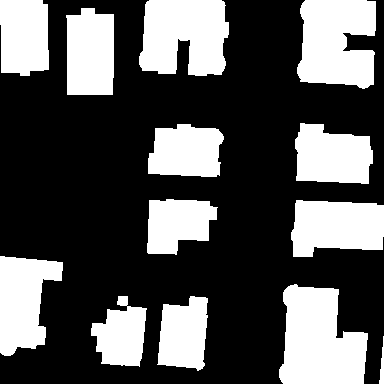}}
		\subfigure[]{\includegraphics[width=3.7cm]{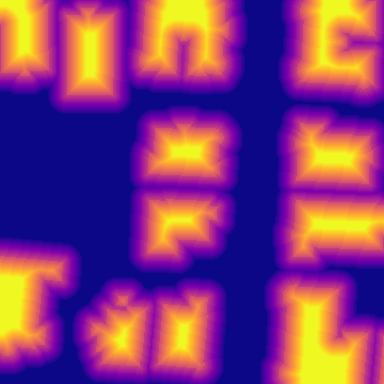}}
		\subfigure[]{\includegraphics[width=3.7cm]{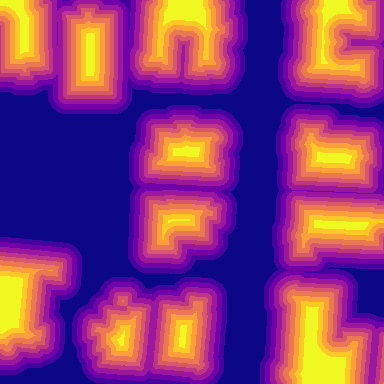}}
	}
	\caption{One example input image with different output representations.
		(a) Satellite Image (RGB),
		(b) Semantic Segmentation Mask,
		(c) Distance Transform,
		(d) Truncated Distance Mask,
		(e) Truncated and Quantized Distance Mask (Best viewed in electronic version)
	}
	\label{fig:rounded_predictions}
\end{figure*}

%%%%%%%%%%%%%%%%%%%%%%%%%
%%
%%
%%		Approach: Cascaded Multi-Task  Network
%%
%%
%%%%%%%%%%%%%%%%%%%%%%%%%
\section{Cascaded Multi-Task  Network} \label{{sec:network}}
In our approach, we use multi-task learning to improve the segmentation predictions of building footprints. The goal is to rely besides the semantic term also a geometric term which incorporates the boundary information of the segmentation mask into a single loss function. We achieve this shared representation of semantic and geometric features by training the network on two different tasks. %By sharing the domain information between complimentary tasks, past work [18, 19] has shown to improve the prediction accuracy of an model compared to training separate models.
We train our segmentation network parameterized by $\theta$ with a set of training
images $x$ along with their ground truth segmentation masks $S$ and corresponding truncated distance class labels $D$ represented by $ {{ (x(n), S(n), D(n)); n =1, 2,...,N }} $. In the following we explain details about the output representation, network architecture and multi-task loss function.

\subsection{Output Representation}

The goal of our multi-task approach is to incorporate besides semantic information about class labels also geometric properties in the network training. Although there are multiple geometric properties which can be extracted such as shape and edge information, we extract on the distance of pixels to boundaries of buildings. Such a representation has the advantages that (1) it can be easily derived from existing segmentation masks by means of the distance transform and (2) neural networks can be easily trained with the representation using existing losses like the mean squared error or the negative log likelihood loss. Using the representation, we bias the network to learn per pixel information about the location of the boundary and capture implicitly geometric properties. We truncate the distance at a given threshold to only incorporate the nearest pixels to the border. Let $Q$ denote the set of pixels on the object boundary and $C_i$ the set of pixels belonging to class $i$. For every pixel $p$ we compute the truncated distance $D(p)$ as 
\begin{equation}
\begin{split}
D(p) = \delta_p min(\min_{\forall q \in Q}  d(p, q), R), \\
\delta_p = \begin{cases}
+1 & \text{if} \quad p \in C_{building} \\
-1 & \text{if} \quad p \notin C_{building}
\end{cases}
\end{split}
\end{equation}

where $d(p,q)$ is the Euclidean distance between pixels $p$ and $q$ and $R$ the truncation threshold. The pixel distances are additionally weighted by the sign function $\delta_p$ to represent whether pixels lie inside or outside the building masks.
The continuous distance values are then uniformly quantized to facilitate training.
%To facilitate training, we additionally transform the continuous representation of truncated distances into a discrete representation by uniformly quantizing distances. 
Similar to Hayder et. al. \cite{hayder2017boundary} we one-hot encode the distance map into a binary vector representation $b(p)$ as:
\begin{equation}
D(p) \sum_{k=1}^{K} r_n b_k(p)  \sum_{k=1}^{K} b_k(p) = 1
\end{equation}
with $r_n$ as distance value corresponding to bin $k$. The $k$ resulting binary pixel-wise maps can be understood as classification maps for each of the $k$th border distance.

\subsection{Enoder-Decoder Network Architecture}
\begin{figure*}[ht!]\label{fig:architecture}
	\centering
	\centerline{
		\includegraphics[width=\textwidth]{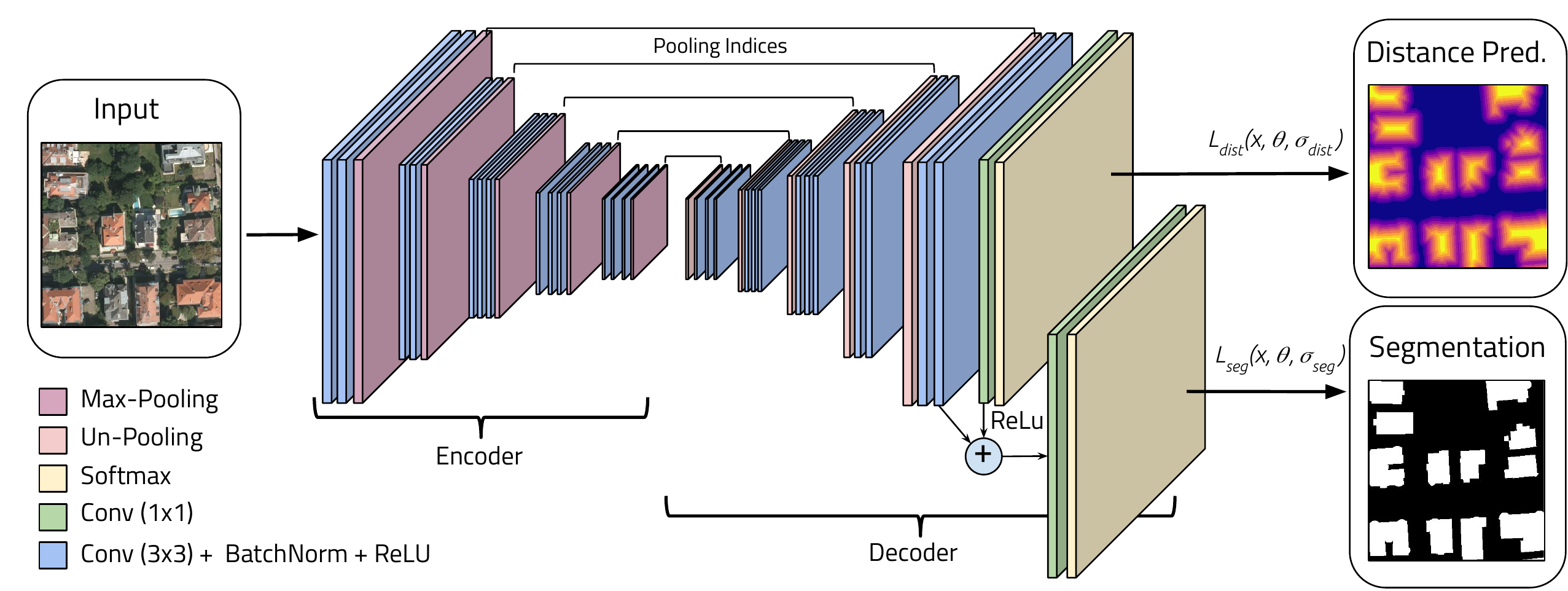}
	}
	\caption{An illustration of the proposed multi-task cascade architecture for semantic segmentation. The encoder is based on the VGG16 architecture. The decoder upsamples its input using the transferred pooling indices from its encoder to densifies the feature maps with multiple successive convolutional layers. The network uses one convolutional layer $H_{dist}$ after the last decoder layer to predict the distance classes. Feature maps produced by $H_{dist}$ and the last layer of the decoder are concatenated and passed to second convolutional layer $H_{seg}$ to compute the final segmentation masks. }
	\label{fig:rounded_predictions}
\end{figure*}

The network in this work is based on the fully convolutional network SegNet \cite{badrinarayanan2015segnet}. SegNet has an encoder-decoder architecture which is commonly used for semantic segmentation. The encoder has the same architecture as VGG16 \cite{simonyan2014very}, consists of 13 convolutional layers of 3x3 convolutions and five layers of 2x2 max pooling. The decoder is a mirrored version of the encoder which uses the pooling indices of the encoder to upsample the feature maps. 
A detailed illustration of the architecture is shown in Fig. \ref{fig:architecture}. We add one convolutional layer $H_{dist}$ to the last layer of the decoder to predict the distance to the border of buildings. The final segmentation mask of building footprints is computed by a second convolutional layer $H_{seg}$. $H_{seg}$ uses the concatenated feature maps of the last decoder layer and the feature maps produced by $H_{dist}$. Thereby the network can leverage semantic properties present in feature maps of the decoder and the geometric properties extracted by $H_{dist}$. Please note, that before the concatenation we pass feature maps of $H_{dist}$ additionally through a ReLU. We finally squash the outputs of $H_{seg}$ and $H_{dist}$ through a softmax layer to get the probabilities for the class labels. 

%We add two convolutional layers Hdist for distance prediction and Hseg for segmentation. Hdist receives the decoder output as input. Hseg receives the concatenation of the decoder output and that of Hdist.

\subsection{Uncertainty Based Multi-Task Loss}

We define the multi-task loss as follows:

\begin{equation} \label{eq:multitaskloss}
	L_{total}(x;\theta) = 	\sum_{i=1}^{T}  \lambda_{i} L_{i}(x;\theta) 
\end{equation}

where $T$ is the number of tasks and $L_{i}$ the corresponding task loss functions to be minimized with respect to the network parameters $\theta$. Each task loss $L_{i}$ is weighted by a scalar $\lambda_{i}$ to model the importance of each task on the combined loss $L_{total}$. The weighting terms $\lambda_{i}$ in the multi-task loss introduce additional hyper-parameters which are usually equated or found through an expensive grid-search. Motivated by Kendall et. al. \cite{kendall2017multi}, we learn the relative task weights $\lambda_{i}$ by taking the uncertainty in the model's prediction for each task into consideration. The aim is to learn a relative task weight depending on the confidence of the individual task prediction. Within the context, we define the multi-loss function $L_{total}$ as a combination of two pixel-wise classification losses. We write the total objective as follows:
\begin{equation}
%\begin{split}
%L_{total}(x;\theta, \sigma_{dist}, \sigma_{seg}) = \lambda_{dist} L_{dist}(x;\theta, %\sigma_{dist})  \\ +  \lambda_{seg} L_{seg}(x;\theta, \sigma_{seg}) 
%\end{split}
\begin{split}
L_{total}(x;\theta, \sigma_{dist}, \sigma_{seg}) = L_{dist}(x;\theta, \sigma_{dist})  \\ 
+ L_{seg}(x;\theta, \sigma_{seg}) 
\end{split}
\end{equation}
where $L_{dist}, L_{seg}$ are the classification loss functions for the prediction of the distance-classes and the segmentation mask with $\sigma_{dist}, \sigma_{seg}$ as corresponding task weights for $\lambda_{i}$.

We represent the likelihood of the model for each classification task as a scaled version of the model output $f(x)$ with the uncertainty $\sigma$ squashed through a softmax function:
\begin{equation}
P(C = 1|x, \theta, \sigma_t) = \frac{exp( \frac{1}{\sigma_t^2} f_c(x))}{ \sum_{c'=1}^{} exp( \frac{1}{\sigma_t^2} f_{c'}(x))}
\end{equation}
Using the negative log likelihood, we express the classification loss with uncertainty as follows:
\begin{equation} \label{eg:class_loss}
\begin{split}
L_t(x, \theta, \sigma_t) = \sum_{c=1}^{C} - C_c log P(C_c=1| x, \theta, \sigma_t) \\
%L_c(x, \theta, c) 
= \sum_{c=1}^{C} - C_c log(exp( \frac{1}{\sigma_t^2} f_c(x))) 
							 + log \sum_{c'=1}^{C}  exp( \frac{1}{\sigma_t^2} f_{c'}(x))
\end{split}
\end{equation}
Applying the same assumption as in \cite{kendall2017multi}:
\begin{equation}
	 \frac{1}{\sigma^2} \sum_{c'}^{} exp( \frac{1}{\sigma^2} f_{c'}(x)) 
	 \approx 
	 (\sum_{c'}^{} exp(f_{c'}(x)))^ \frac{1}{\sigma^2}
\end{equation}
allows to simplify Eq. \ref{eg:class_loss} to:
\begin{equation}\label{eq:loss_approx}
\begin{split}
L_t(x, \theta, \sigma_t) \approx \frac{1}{\sigma_t^2} \sum_{c=1}^{C} - C_c log P(C_c=1| x, \theta) + log(\sigma_t^2)
\end{split}
\end{equation}
We use the approximated form of Eq. \ref{eq:loss_approx} in both classification tasks $L_{dist}$ and $ L_{seg}$ for the prediction of segmentation classes and distance classes respectively. It is important to note, that for numerical stability, we trained the network to predict $log(\sigma_i^2)$ instead of $\sigma_i^2$. All network parameters and the uncertainty tasks weights are optimized with stochastic gradient descent (SGD).

\begin{table*}[ht!]
	\centering
	\caption{This table shows the evaluation results of the same network using single task vs. multi-task. Both network using the multi-task loss outperform the single task predictions. The uncertainty based task weights lead to an further improvement and achieve overall the best results.}
	\label{tab:results_per_location}
\begin{tabular}{cccccccc}
	& \multicolumn{1}{c|}{}                                                   & \multicolumn{1}{c|}{Austin}                                                         & \multicolumn{1}{c|}{Chicago}                                                        & \multicolumn{1}{c|}{Kitsap Co.}                                                     & \multicolumn{1}{c|}{West Tyrol}                                            & \multicolumn{1}{c|}{Vienna}                                                         & Overall                                                        \\ \hline
	\begin{tabular}[c]{@{}c@{}}FCN + MLP\\ (Baseline)\end{tabular}                            & \multicolumn{1}{c|}{\begin{tabular}[c]{@{}c@{}}IoU\\ Acc.\end{tabular}} & \multicolumn{1}{c|}{\begin{tabular}[c]{@{}c@{}}61.20\\ 94.20\end{tabular}}          & \multicolumn{1}{c|}{\begin{tabular}[c]{@{}c@{}}61.30\\ 90.43\end{tabular}}          & \multicolumn{1}{c|}{\begin{tabular}[c]{@{}c@{}}51.50\\ 98.92\end{tabular}}          & \multicolumn{1}{c|}{\begin{tabular}[c]{@{}c@{}}57.95\\ 96.66\end{tabular}} & \multicolumn{1}{c|}{\begin{tabular}[c]{@{}c@{}}72.13\\ 91.87\end{tabular}}          & \begin{tabular}[c]{@{}c@{}}64.67\\ 94.42\end{tabular}          \\ \hline
	\begin{tabular}[c]{@{}c@{}}SegNet (Single-Loss) \\ NLL-Loss for Seg. Classes\end{tabular} & \multicolumn{1}{c|}{\begin{tabular}[c]{@{}c@{}}IoU\\ Acc.\end{tabular}} & \multicolumn{1}{c|}{\begin{tabular}[c]{@{}c@{}}74.81\\ 92.52\end{tabular}}          & \multicolumn{1}{c|}{\begin{tabular}[c]{@{}c@{}}52.83\\ 98.65\end{tabular}}          & \multicolumn{1}{c|}{\begin{tabular}[c]{@{}c@{}}68.06\\ 97.28\end{tabular}}          & \multicolumn{1}{c|}{\begin{tabular}[c]{@{}c@{}}65.68\\ 91.36\end{tabular}} & \multicolumn{1}{c|}{\begin{tabular}[c]{@{}c@{}}72.90\\ 96.04\end{tabular}}          & \begin{tabular}[c]{@{}c@{}}70.14\\ 95.17\end{tabular}          \\ \hline
	\begin{tabular}[c]{@{}c@{}}SegNet (Single-Loss)\\ NLL-Loss for Dist. Classes\end{tabular} & \multicolumn{1}{c|}{\begin{tabular}[c]{@{}c@{}}IoU\\ Acc.\end{tabular}} & \multicolumn{1}{c|}{\begin{tabular}[c]{@{}c@{}}76.49\\ 93.12\end{tabular}}          & \multicolumn{1}{c|}{\begin{tabular}[c]{@{}c@{}}66.77\\ 99.24\end{tabular}}          & \multicolumn{1}{c|}{\begin{tabular}[c]{@{}c@{}}72.69\\ 97.79\end{tabular}}          & \multicolumn{1}{c|}{\begin{tabular}[c]{@{}c@{}}66.35\\ 91.58\end{tabular}} & \multicolumn{1}{c|}{\begin{tabular}[c]{@{}c@{}}76.25\\ 96.55\end{tabular}}          & \begin{tabular}[c]{@{}c@{}}72.57\\ 95.66\end{tabular}          \\ \hline
	\begin{tabular}[c]{@{}c@{}}SegNet + MultiTask-Loss\\ (Equally Weighted)\end{tabular}      & \multicolumn{1}{c|}{\begin{tabular}[c]{@{}c@{}}IoU\\ Acc.\end{tabular}} & \multicolumn{1}{c|}{\begin{tabular}[c]{@{}c@{}}76.22\\ 93.03\end{tabular}}          & \multicolumn{1}{c|}{\begin{tabular}[c]{@{}c@{}}66.64\\ 99.24\end{tabular}}          & \multicolumn{1}{c|}{\begin{tabular}[c]{@{}c@{}}71.70\\ 97.71\end{tabular}}          & \multicolumn{1}{c|}{\begin{tabular}[c]{@{}c@{}}\textbf{67.03}\\ 91.66\end{tabular}} & \multicolumn{1}{c|}{\begin{tabular}[c]{@{}c@{}}\textbf{76.68}\\ 96.60\end{tabular}}          & \begin{tabular}[c]{@{}c@{}}72.65\\ 95.65\end{tabular}          \\ \hline
	\begin{tabular}[c]{@{}c@{}}SegNet + MultiTask-Loss\\ (Uncertainty Weighted)\end{tabular}  & \multicolumn{1}{c|}{\begin{tabular}[c]{@{}c@{}}IoU\\ Acc.\end{tabular}} & \multicolumn{1}{c|}{\textbf{\begin{tabular}[c]{@{}c@{}}76.76\\ 93.21\end{tabular}}} & \multicolumn{1}{c|}{\textbf{\begin{tabular}[c]{@{}c@{}}67.06\\ 99.25\end{tabular}}} & \multicolumn{1}{c|}{\textbf{\begin{tabular}[c]{@{}c@{}}73.30\\ 97.84\end{tabular}}} & \multicolumn{1}{c|}{\begin{tabular}[c]{@{}c@{}}66.91\\ \textbf{91.71}\end{tabular}} & \multicolumn{1}{c|}{\textbf{\begin{tabular}[c]{@{}c@{}}76.68\\ 96.61\end{tabular}}} & \textbf{\begin{tabular}[c]{@{}c@{}}73.00\\ 95.73\end{tabular}} \\ \hline
	\multicolumn{1}{l}{}                                                                      & \multicolumn{1}{l}{}                                                    & \multicolumn{1}{l}{}                                                                & \multicolumn{1}{l}{}                                                                & \multicolumn{1}{l}{}                                                                & \multicolumn{1}{l}{}                                                       & \multicolumn{1}{l}{}                                                                & \multicolumn{1}{l}{}                                           \\
	\multicolumn{1}{l}{}                                                                      & \multicolumn{1}{l}{}                                                    & \multicolumn{1}{l}{}                                                                & \multicolumn{1}{l}{}                                                                & \multicolumn{1}{l}{}                                                                & \multicolumn{1}{l}{}                                                       & \multicolumn{1}{l}{}                                                                & \multicolumn{1}{l}{}                                          
\end{tabular}
\end{table*}

\section{Experimental Results}

\subsection{Inria Aerial Image Labeling Dataset}

The Inria Aerial Image Labeling Dataset \cite{maggiori2017can} is comprised of 360 ortho-rectified aerial RGB images at 0.3m spatial resolution. The satellite scenes have tiles of size 5000 x 5000 px, thus covering a surface of 1500 x 1500m per tile. The images comprise ten cities and an overall area of 810 sq. km. The images convey dissimilar urban settlements, ranging from densely populated areas (e.g., San Francisco’s financial district) to alpine towns (e.g,. Linz in Austrian Tyrol). Ground-truth data is provided for the two semantic classes \textit{building} and \textit{non-building}. The ground truth is only provided for the training set with covers five cities. For comparability, we split the dataset as described by Maggiori et al. \cite{maggiori2017can} (image 1 to 5 of each location for validation, 6 to 36 for training).
 
 \subsection{Evaluation Metrics}
 We evaluate our approach in the following experiments with two metrics. The first one is the Intersection over Union (IoU) for the positive building class. This is the number of pixels labeled as building in the prediction and the ground truth, divided by the number of pixels labeled as pixel in the prediction or the ground truth. As second metric, we report accuracy, the percentage of correctly classified pixels.
 
%\subsection{Experiments and Discussion}

\subsection{Importance of a Deeper Encoder-Decoder-Architecture} \label{experiment_1}
In the first experiment, we analyze the importance of the encoder and decoder architecture. Unlike the past work \cite{maggiori2016high, maggiori2017can}, we used the deeper network based on VGG16 \cite{simonyan2014very} as encoder and evaluate different decoder architectures. In our comparison we train the following networks:
\begin{enumerate}
	\item a FCN \cite{shelhamer2017fully} which uses an up-sampling layer and convolutional layer as decoder
	\item a SegNet \cite{badrinarayanan2015segnet} which attaches a reversed VGG16 as decoder to the encoder
	\item the combination of FCN + MLP as introduced in \cite{maggiori2017can} which up-samples and concatenates all feature maps of the FCN encoder and uses a MLP to reduce the feature maps to class predictions. This approach achieves currently the highest accuracy on the dataset.
\end{enumerate}
Where applicable, we initialize the weights of the encoder with the weights of a VGG16 \cite{simonyan2014very} model pre-trained via ImageNet \cite{russakovsky2015imagenet}. All networks are then trained with SGD using a learning rate of 0.01, weight decay of 0.0005 and momentum of 0.9. We use the negative log likelihood loss on the segmentation class labels, reduce the learning rate every 25,000 iterations by the factor 0.1 and stop the training after 200,000 iterations. We extract 10 mini-batches from each satellite image and randomly crop four patches of size 384 x 384 pixels for each mini-batch from the satellite scenes. We apply randomly flipping in vertical and horizontal directions as data augmentation. 
Table \ref{tab:tab1} shows the results of the different architectures on the validation set. The following observations can be made from the table: (1) Due to the deeper architecture of the encoder all architectures outperform previous state-of-the-art approaches on the dataset. This indicates that features learned by the networks proposed in the past were not expressive enough for the segmentation task. (2) When comparing SegNet against the proposed method \cite{maggiori2017can} we do not observe an improvement. This indicates that the SegNet decoder produces better results as compared to the SegNet + MLP combination. (3) The different architectures show that the decoder plays a crucial role for the semantic segmentation task. While the FCN achieves an IoU of 66.21\% for the building class, we see an improvement of 3.9\% (against the FCN) when using the same encoder but the more complex decoder as in SegNet.

\begin{table}
	\centering
	\caption{This table lists the prediction accuracies for the segmentation masks in \ref{experiment_1} on the validation set. All networks use the same encoder but different decoder types. All VGG16-based models outperform state-of-the-art, while SegNet improves the IoU by more than 5\%.}
	\label{tab:tab1}
	\begin{tabular}{l|c|c}
		& mean IoU & Acc. (Pixel) \\ \hline
		Baseline FCN \cite{maggiori2017can}                  & 53.82\%        			 & 92.79 \%                \\ \hline
		Baseline FCN + MLP\cite{maggiori2017can}        & 64.67\%        			& 94.42 \%                \\ \hline
		FCN (VGG16 encoder)				                  						  & 66.21\%        			   & 94.54 \%                \\ \hline
		FCN + MLP (VGG16 encoder)							                 				& 68.17\%        			 & 94.95 \%                \\ \hline
		SegNet (VGG16 encoder)								                  					& \textbf{70.14}\%     & \textbf{95.17} \%                \\ \hline
	\end{tabular}
\end{table}

%In the following experiments we evaluate our network architecture and the proposed multi-task loss more detailed:
%details about R, number of bins

\subsection{Importance of Distance Prediction}
In this section, we evaluate the advantage of predicting distance classes to boundaries using a single loss function. As baseline we take SegNet from the previous experiment which was trained with the NLL loss on the semantic segmentation classes and achieved the best results of 70.14\% IoU for the building class. 
We modify this network such that we remove the $H_{seg}$ and attach $H_{dist}$ as shown in \ref{fig:architecture}. As output representation we use the truncated and quantized distance mask, setting the truncation threshold $R$=20 and the number of bins $K$=10. We train the network as in the previous experiment with SGD and let the network predict distance classes to boundaries. To get the final segmentation mask from the distance predictions, we threshold all distances above five to only get the pixels inside the buildings. The results for this approach are illustrated in Table \ref{tab:results_per_location} and show that by relying on boundary information, we can improve the overall IoU for SegNet by about 2.4\%.

\begin{figure*}[th!]
	\centerline{
		\includegraphics[width=2.8cm]{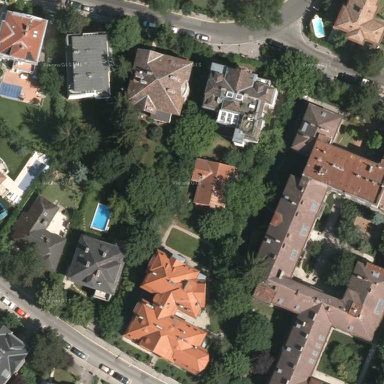}
		\includegraphics[width=2.8cm]{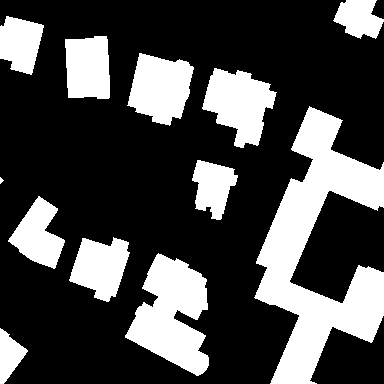}
		\includegraphics[width=2.8cm]{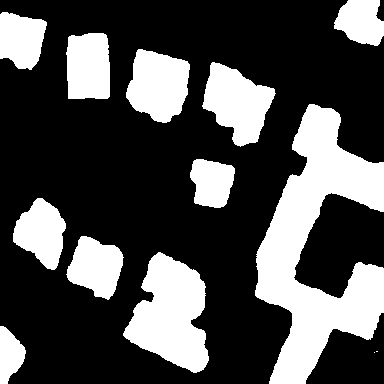}
		\includegraphics[width=2.8cm]{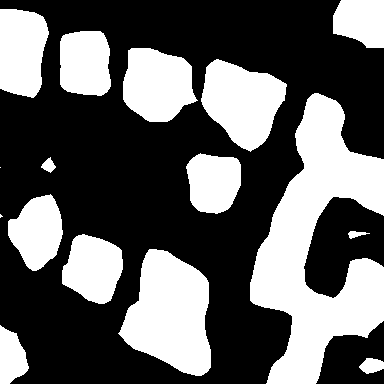}
		\includegraphics[width=2.8cm]{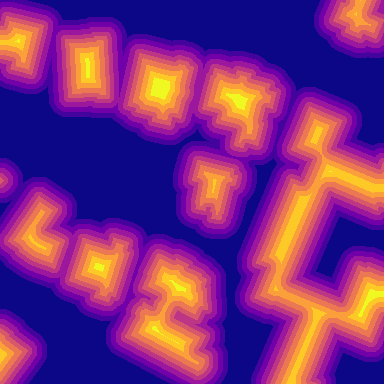}
		\includegraphics[width=2.8cm]{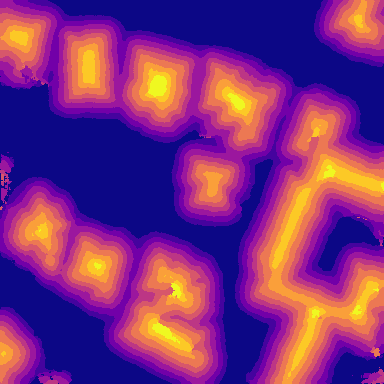}
	}
\end{figure*}
\begin{figure*}[th!]
	\vspace{-11pt}
	\centerline{
		\subfigure[]{ \includegraphics[width=2.8cm]{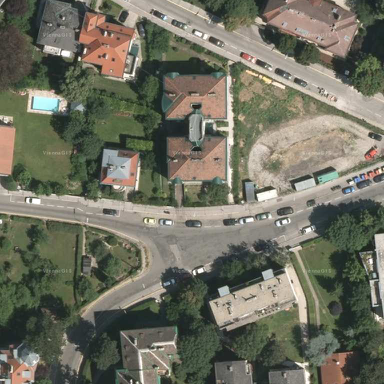}}
		\subfigure[]{\includegraphics[width=2.8cm]{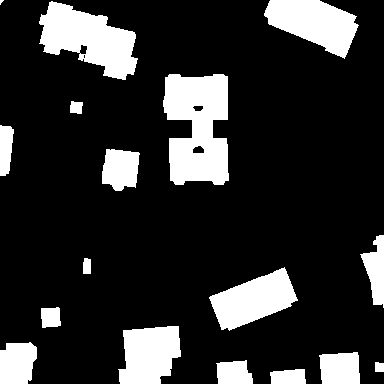}}
		\subfigure[]{\includegraphics[width=2.8cm]{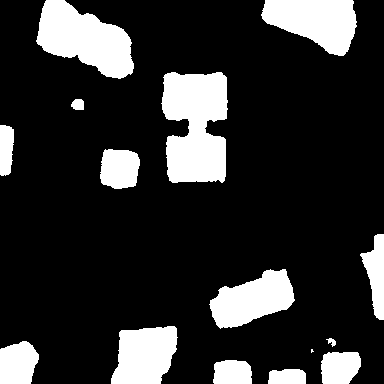}}
		\subfigure[]{\includegraphics[width=2.8cm]{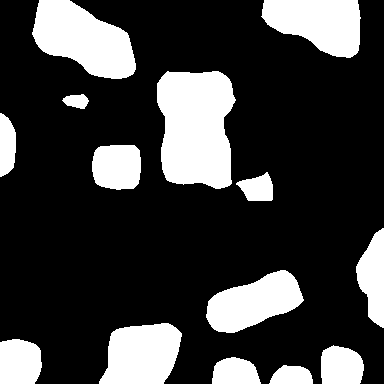}}
		\subfigure[]{\includegraphics[width=2.8cm]{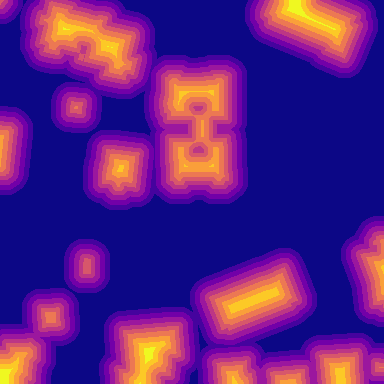}}
		\subfigure[]{\includegraphics[width=2.8cm]{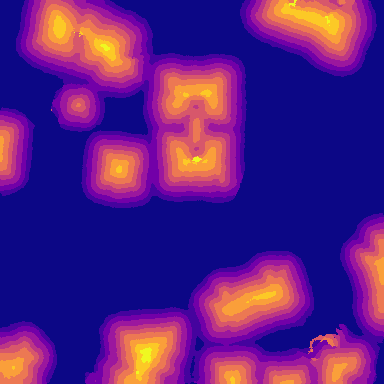}}
	}
	\caption{Two different locations, column-wise: (a) satellite images in RGB, (b) ground truth masks for the building footprints, (c) segmentation predictions by our proposed multi-task network, (d) segmentation predictions by an FCN \cite{shelhamer2017fully}, (e) ground truth masks for distance classes and (f) predicted distance classes. It can be seen that our approach produces less ``blobby" predictions with sharper edges compared to the FCN. (Best viewed in electronic version)}
	\label{fig:final_predictions}
\end{figure*}

\subsection{Importance of Uncertainty Based Multi-Task Learning}

In the last experiment, we show the advantage the multi-task loss combining boundary and semantic information to improve the segmentation result. We initialized the network shown in Fig. \ref{fig:architecture} with the network-weights from the previous experiment and retrain it with the uncertainty based multi-task loss using Eq. \ref{eq:multitaskloss}. The network is trained with SGD using an initial learning rate of 0.001, weight decay of 0.0005 and momentum of 0.9. To evaluate the influence on the uncertainty weights we additionally train the same network using the multi-task loss but set the importance factors $\lambda_i$ of both tasks to one.
The results on Table \ref{tab:results_per_location} illustrate that the uncertainty task loss achieves per location and overall on both evaluation metrics the best results. When using the equally weighted multi-task loss, the overall accuracy is better compared to both single loss tasks but worse than the uncertainty weighted multi task loss. The final prediction results for segmentation masks and distance classes are illustrated in Fig. \ref{fig:final_predictions}.

\section{Conclusion}
In this paper, we addressed the problem of incorporating geometric information into the internal representation of deep neural networks. We therefore focused on semantic segmentation of building footprints from high resolution satellite imagery and showed how boundary information of segmentation masks can be leveraged using a multi-task loss. Our proposed approach outperforms recent methods on the Inria Aerial Image Labeling Dataset significantly by 8.3\% which shows the effectiveness of our work. Building upon this work, we plan to extend our multi-task network with further geometric cues and to multiple classes to preserve semantic boundaries. In this context we also plan to make the step from semantic segmentation towards instance segmentation.

% you can choose not to have a title for an appendix
% if you want by leaving the argument blank
% use section* for acknowledgment
\section*{Acknowledgment}
The authors would like to thank NVIDIA for support within the NVAIL program. Additionally, this work was supported BMBF project MOM (Grant 01IW15002).

\bibliographystyle{IEEEtran}
\bibliography{building_extraction}

\end{document}